\documentclass[Afour,times,sageh,doublespace]{sagej}

\usepackage{moreverb,url}
\usepackage{mathtools}
\usepackage{float}

\usepackage{multirow}
\usepackage{placeins}
\usepackage{multicol}

\usepackage{dirtree}

\usepackage[colorlinks,bookmarksopen,bookmarksnumbered,citecolor=red,urlcolor=red]{hyperref}

\newcommand\BibTeX{{\rmfamily B\kern-.05em \textsc{i\kern-.025em b}\kern-.08em
T\kern-.1667em\lower.7ex\hbox{E}\kern-.125emX}}

\def\volumeyear{2023}

\begin{document}

\runninghead{Randall and Treibitz}

\title{FLSea: Underwater Visual-Inertial and Stereo-Vision Forward-Looking Datasets}

\author{Yelena Randall\affilnum{1} and Tali Treibitz\affilnum{1}}

\affiliation{\affilnum{1}ViSEAon Marine Imaging Lab,
Department of Marine Technologies,
University of Haifa,
Haifa, Israel\\
\\
The VI dataset can be downloaded from \url{https://www.kaggle.com/datasets/viseaonlab/flsea-vi}.
}

\corrauth{Yelena Randall, ViSEAon Marine Imaging Lab,
Department of Marine Technologies,
University of Haifa,
Haifa, Israel}

\email{y4randall@gmail.com}

\begin{abstract}
Visibility underwater is challenging, and degrades as the distance between the subject and camera increases. That is why underwater  computer vision tasks in the forward-looking direction are more difficult.  We have collected underwater forward-looking stereo-vision and visual-inertial image sets with two underwater imaging platforms, a stereo camera rig and an ROV in the Mediterranean and Red Sea. To our knowledge there is only one other public dataset in the underwater environment with this camera-sensor orientation. These datasets are critical for the development of several underwater applications, including autonomous obstacle avoidance, visual odometry, 3D tracking, Simultaneous Localization and Mapping (SLAM) and depth estimation through deep learning. The stereo datasets include synchronized stereo images in dynamic underwater environments with objects of known-size. The visual-inertial datasets contain monocular images and IMU measurements, aligned with millisecond resolution timestamps and objects of known size which were placed in the scene. Both sensor configurations allow for scale estimation, with the calibrated baseline in the stereo setup and the IMU in the visual-inertial setup. Ground truth depth maps were created offline for both dataset types using a commercial photogrammetry software (Agisoft Metashape). The ground truth is validated with multiple known measurements placed throughout the imaged environment. There are 5 stereo and 8 visual-inertial datasets in total, each containing thousands of images, with a range of different underwater visibility and ambient light conditions, natural and man-made structures and dynamic camera motions. The forward-looking orientation of the camera makes these datasets unique and ideal for testing underwater obstacle-avoidance algorithms and for navigation close to the seafloor in dynamic environments. With our datasets, we hope to encourage the advancement of autonomous functionality for underwater vehicles in dynamic and/or shallow water environments.

\end{abstract}


\maketitle

\section{Introduction}
Autonomous underwater vehicles (AUVs) are designed with the goal of operating untethered, navigating and performing tasks without user input. They use a combination of acoustic, inertial and visual sensors to sense the world around them and act accordingly. Typically, AUVs operate from a downward-looking field of view, in the middle of the water column~\cite{aqualoc}. This allows them to collect data from above, and stay at a safe distance from obstacles. In this case, the obstacle avoidance protocol can be  based purely on bathymetry, where the AUV's only requirement is to stay far enough above any obstacles below~\cite{sparus_ii}. In our work, we consider an alternate scenario, where the AUV observes in the forward-looking field-of-view. This setup would allow the vehicle to see what's ahead of it and operate in complex, shallow-water environments, facing obstacles head-on and collecting data much closer to the subject of interest. For this, scaled 3D reconstruction in real-time is required. \\
The sensor best fitted for this task is an optical camera for multiple reasons. It provides data at a close range to the subject, it is affordable compared to acoustic sensors, it is lightweight and small, and is widely available. We choose a monocular camera over a stereo camera configuration because on a small system as we imagine, a stereo setup with the required baseline could be cumbersome, and could hinder the vehicle's ability to be agile. With a visible camera we get the advantage of high resolution data at relatively high frame rates. However, there are some challenges of using a camera for underwater 3D reconstruction.  In water, light is attenuated as a function of imaging range and scattered by particles suspended in the medium between the camera and subject~\cite{derya}. This often results in an image with low contrast and in which details are veiled by scattered light (backscatter)~\cite{jaffe}. \\ 
Outside of the underwater domain, there has been much development towards the monocular vision 3D scene reconstruction problem. Structure from motion (SFM), simultaneous localization and mapping (SLAM), and 3D reconstruction with deep learning are a few of the well-known solutions. What all of these methods have in common is the unknown scale issue, meaning that metric scale cannot be derived from monocular images alone. In other words, without a sensor or external cue which provides scaled measurements, scene depth and camera pose are estimated at an arbitrary scale. SFM is the computer vision solution to reconstructing 3D information from multiple images~\cite{colmap}. SFM utilizes camera motion in an assumed stationary world to build a 3D sparse depth map of the scene as well as relative camera pose~\cite{Hartley2004}. It is performed offline. There are a few groups which have built the foundation for SFM~\cite{sfm_og, sfm_uncalibrated},  as well as groups that have expanded on this research to large-scale and urban image sets~\cite{rome_in_a_day, vacation_sfm} and groups that have developed well-known open source and commercial SFM software~\cite{colmap, opensfm, openmvg, pixel_perfect, agisoft}. SLAM, which uses the same principle as SFM, is the solution developed by roboticists to aid with navigation~\cite{orb-slam}. \\
The goal of SLAM is to autonomously navigate through an unknown environment. Therefore, unlike SFM, it works in real-time and tracks camera motion while building a map of the scene simultaneously. SLAM can correct for error accumulated over time by revisiting previously seen locations, a practice known as loop closure. It is especially useful for GPS-limited places and is preferred over odometry-based tracking methods as it can correct for drift over time. There are many different flavors of SLAM, that use different techniques - feature-based or direct, produce different outputs - sparse or dense maps and use different sensor combinations - monocular vision, stereo vision, RGB-D, LIDAR, etc. Considering our desired use-case, we will mention several well-known monocular visual methods such as~\cite{orb-slam3, lsd-slam, dso, svo}. \\
Visual-Inertial odometry (VIO) is another monocular camera based algorithm developed by roboticists which solves for the scale issue by adding an inertial measurement unit (IMU)~\cite{Rosinol20icra-Kimera, r-vio}. VIO is an algorithm used for localization, which estimates the robot state using a monocular camera and an IMU. With the addition of an inertial sensor, VIO makes it possible to estimate the camera pose in metric scale~\cite{vio}. However, building a map of the scene is not the main goal of a pure VIO algorithm and therefore does not consider using loop closure to correct for error.  But, it is becoming more and more common to combine SLAM and VIO because it solves the scale issue in monocular SLAM, only with the addition of an IMU~\cite{orb-slam3, rovio, maplab, vins-fusion}. An IMU is an affordable small sensor and standard on a robotic system. Visual-Inertial SLAM outputs a metric scaled map of the scene, and can correct for error over time using loop closure~\cite{mur-artal}. \\
Outside of the robotics realm, deep learning is becoming used for depth estimation from monocular images~\cite{megadepth, leres, binsformer, newcrfs, glpdepth, pixelformer} but often not designed for real-time or embedded platforms except for~\cite{fastdepth}. The output of a monocular deep learning network is a dense depth map with unknown scale. The standard training for monocular depth estimation networks is on images and ground truth depth maps.\\
These algorithms have had plenty of success for a variety of robotics tasks on-land and there are multiple benchmark datasets used for development such as KITTI, EUROC and TUM RGB-D, to name a few~\cite{kitti, euroc, tum-rgb-d}. While we do see success on land, there are still not many groups utilizing this technology in this sensor configuration underwater. There are a few foreseeable challenges in creating an underwater system which relies on vision for perception and navigation. Besides the effects of the underwater environment on image quality, the unstructured nature of underwater environments will also pose a challenge for algorithms which are designed and normally tested on structured, man-made environments. But, in order to move past these challenges, underwater forward-looking visual data is necessary for development and testing. This data that we speak of is next to non-existent. Collecting data of this type is time-consuming, logistically complicated and costly. For underwater autonomous tasks as we imagine it is necessary to have forward-looking underwater images at a high frame rate, meaning above 10 fps, with scale information for testing and training, and ground truth for validation and supervision. Underwater data is challenging to acquire, and without access to publicly available data, the progress of underwater 3D image reconstruction in the forward-looking view is limited. There are a few downward-looking underwater visual datasets available, the underwater inspection and intervention dataset~\cite{inspection_intervention_dataset}, AQUALOC: An underwater dataset for visual–inertial–pressure localization~\cite{aqualoc} and the underwater caves sonar data set~\cite{underwater_caves}. \\
To our knowledge, there is only one other publicly available forward-looking visual dataset, which contains stereo images and IMU measurements, but does not contain ground truth~\cite{sCarolina}.
One major challenge underwater is that we do not have access to some of the standard methods for collecting ground truth such as LIDAR scanning. Our solution is to use a commercial SFM-based software called Agisoft Metashape~\cite{agisoft} to create dense depth maps for each image frame in the datasets using objects of known size to validate the result. In the visual-inertial sets, one of the objects of known size is used to scale the ground truth model. \\
Our collection of datasets, titled FLSea, contains two types of forward-looking visual data, collected on a stereo setup and a monocular visual-inertial setup. The stereo datasets consist of synchronized pairs of images, with objects of known size placed in the scene, intrinsic and extrinsic calibrations and ground truth depth maps for each frame. The visual-inertial datasets consist of monocular images and IMU readings, with millisecond resolution timestamps, objects of known size placed in the scene, intrinsic and extrinsic calibrations and ground truth depth maps.

These datasets, together with the ground-truth that we meticulously generated, can be used in evaluating and training VIO, VI-SLAM, Stereo SLAM and monocular depth reconstruction algorithms. Having access to public visual data advances the development of any visual task irregardless of domain, and especially in the underwater domain where data of this type is lacking. We hope that it will serve as a benchmark dataset for underwater forward-looking tasks.

\section{Sensor setup}
The datasets were collected on two different imaging platforms, a diver-held stereo rig and the BlueROV2~\cite{bluerov2}, the visual-inertial platform. 

\subsection{Stereo}
The stereo rig is comprised of two Nikon D810s, in Hugyfot underwater housings which are mounted on a diver-held rig, with a fixed baseline. The housings are pressure rated to 100~m. The housing is equipped with a dome port, to minimize distortion that occurs by refraction at the interface between water and air. The two cameras are hardware synchronized, using a custom synchronization cable that sends a signal to the other camera when either one of the shutter buttons are pressed. The stereo images resolution is 1280 $\times$ 720 and captured at 10~Hz in video mode. Because we are capturing in video mode, there is only one joint hardware trigger, at the start of the capture, which allows for a higher frame rate, but makes it so there is not a hardware trigger for each image frame. In order to ensure that there was no time-lag between left and right camera frames in this mode, a simple test was performed. We filmed a stopwatch with millisecond units, extracted the text from each frame, and confirmed that there was no time difference between any image frame. This ensures synchronization in a millisecond resolution, which is enough with our swimming pace.

\begin{table}[h]
\small\sf\centering
\caption{Technical specifications of the stereo setup.\label{stereo}}
\begin{tabular}{p{2.5cm} p{4cm}} 
\toprule
Camera ($\times$2) &\\
\midrule
Camera model & Nikon D810\\
Sensor type & full-frame CMOS sensor\\
Sensor size & 35.9×24~mm\\
Focal length & 35~mm\\
Opening angle & 54.3$^{\circ}$\\
Resolution & 1280 $\times$ 720\\
Frequency & 10~Hz\\
\bottomrule
\end{tabular}\\[10pt]
\begin{tabular}{p{2.5cm} p{4cm}}
\toprule
\multicolumn{2}{l}{Underwater housing ($\times$2)} \\
\midrule
Housing model & Hugyfot HFN-D810\\ 
Depth rating & 100~m\\
Port & Dome port\\
\bottomrule
\end{tabular}
\end{table}

\begin{figure}
    \centering
    \includegraphics[width=\linewidth]{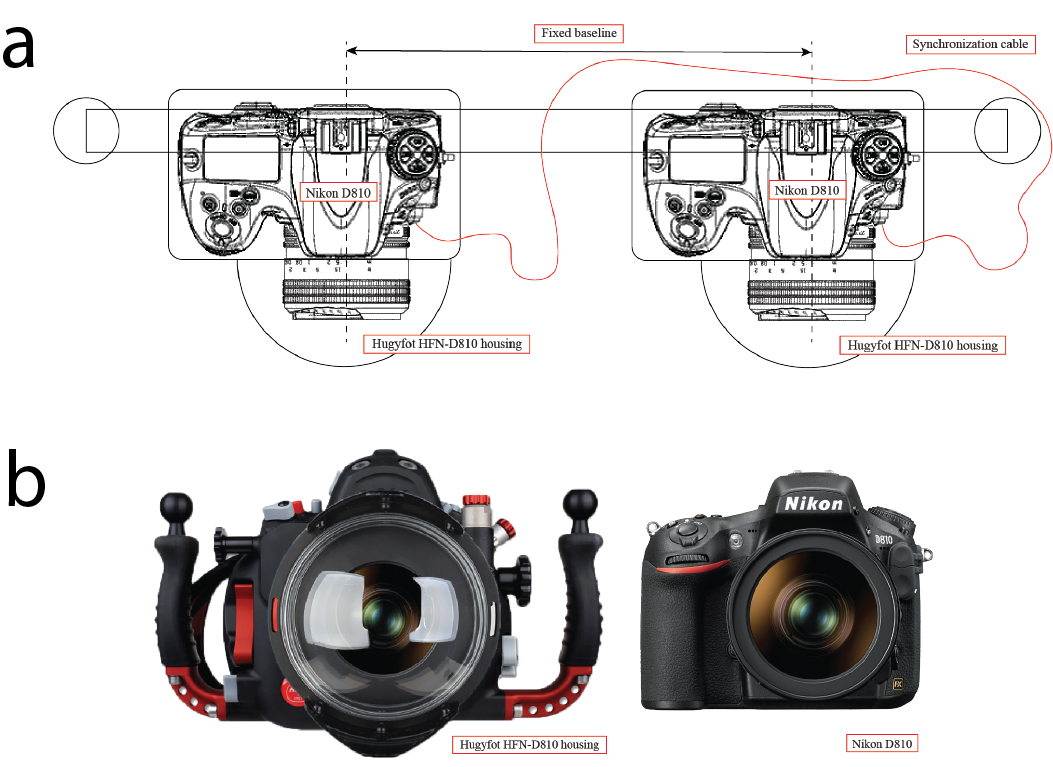}
    \caption{a) Stereo setup diagram with the synchronization cable and outline of the static jig (fixed baseline) and underwater housings sketched, camera sketches~\cite{nikon_sketch}, b) Stereo setup components. [Left]~Underwater housing~\cite{hugyfot}. [Right]~A Nikon SLR camera~\cite{nikond810}.}
\end{figure}

\subsection{Visual-Inertial}
We used the BlueROV2~\cite{bluerov2} as our visual-inertial system. It contains open-source electronics and software, and was customized in our lab. It is equipped with an IDS camera and a VectorNav IMU, running on an Nvidia Jetson Nano as the embedded processor in a housing pressure rated to 100~m. The housing is equipped with a dome part. The images are of resolution 968 $\times$ 608, captured at 10~Hz. There is no hardware synchronization between the IMU and the camera but both save messages with millisecond resolution timestamps. The capture rate of the IMU is 10 times more than that of the camera, making the time delay between any one camera frame and the nearest IMU reading  negligible. 

\begin{table}[h]
\small\sf\centering
\caption{Technical specifications of the visual-inertial setup.\label{vi}}
\begin{tabular}{p{2.5cm} p{4.2cm}} 
\toprule
Camera &\\
\midrule
Camera model & iDS camera (UI-3260CP Rev. 2)\\
Sensor type & Sony IMX249, 1/1.2" CMOS\\
Sensor size & 2.35 Megapixels\\
Focal length & 4~mm\\
Opening angle & 80$^{\circ}$\\
Resolution & 968 $\times$ 608\\
Frequency & 10~Hz\\
\bottomrule
\end{tabular}\\[10pt]
\begin{tabular}{p{2.5cm} p{4.2cm}}
\toprule
\multicolumn{2}{l}{IMU} \\
\midrule
IMU model & VectorNav (VN-100)\\ 
Frequency & 100~Hz\\
\bottomrule
\end{tabular}\\[10pt]
\begin{tabular}{p{2.5cm} p{4.2cm}}
\toprule
\multicolumn{2}{l}{Computer} \\
\midrule
Processor model & Nvidia Jetson Nano\\ 
\bottomrule
\end{tabular}\\[10pt]
\begin{tabular}{p{2.5cm} p{4.2cm}}
\toprule
\multicolumn{2}{l}{Underwater housing} \\
\midrule
Housing model & Blue robotics 4'' tube\\ 
Depth rating & 100~m\\
Port & Blue robotics 4'' dome port\\
\bottomrule
\end{tabular}\\
\end{table}

\begin{figure}
    \centering
    \includegraphics[width=\linewidth]{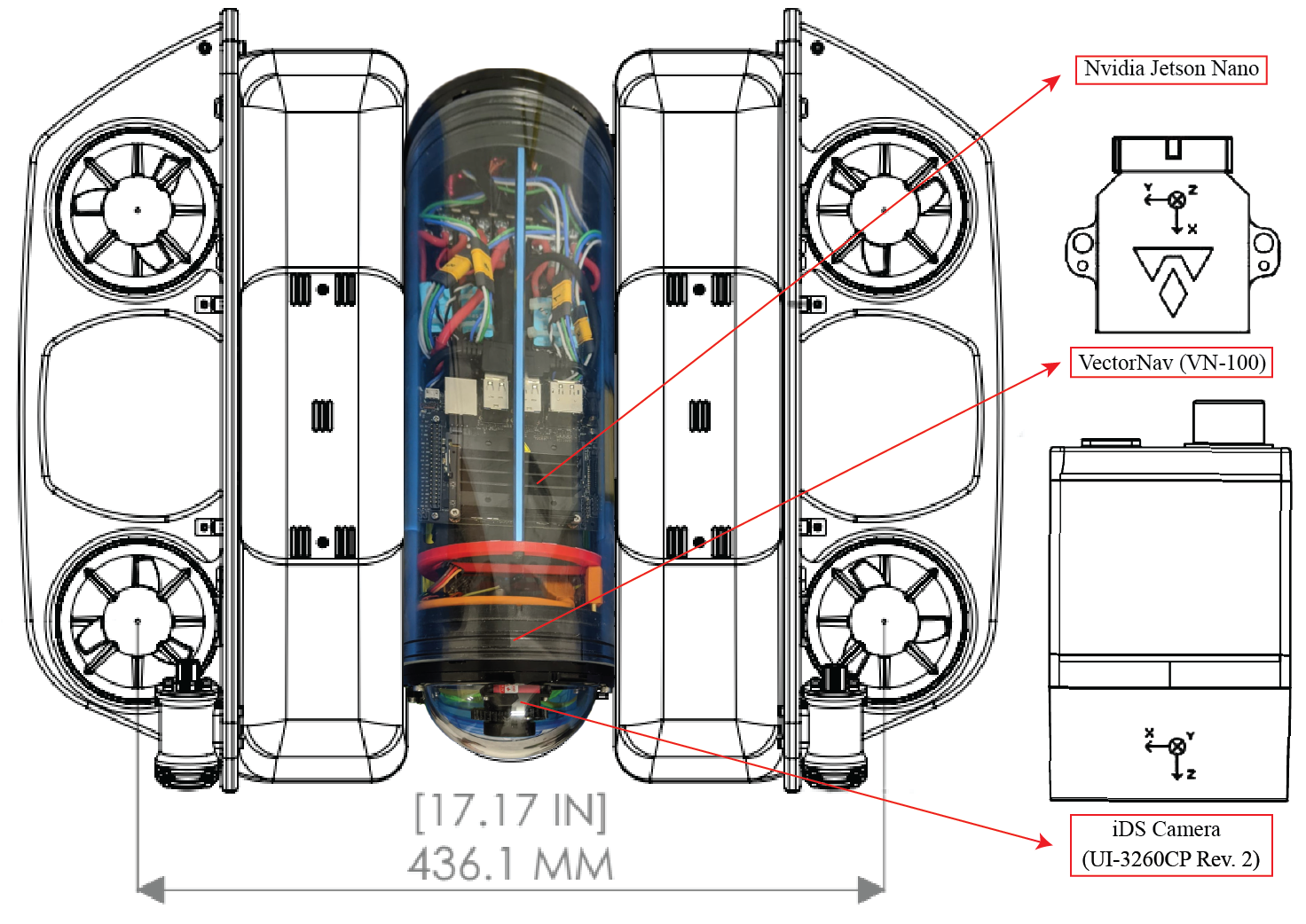}
    \caption{The BlueROV 2 in the four-thruster configuration. The notable sensors here are the VN-100 IMU~\cite{vn-100} and the iDS camera \cite{ids-cam}. Blueprint from~\cite{bluerov2}, center tube is the actual payload tube from our BlueROV2 setup.}
\end{figure}

\subsection{Calibration}
Each system has been calibrated to find intrinsic and extrinsic parameters. The intrinsic parameters of a camera are the focal length, optical center and distortion coefficients. The intrinsic parameters are expressed by a $3 \times 3$ matrix, $K$, which holds the focal length $f_x,f_y$ and optical center $c_x,c_y$
\begin{equation}
    K = \begin{bmatrix}
            f_x&0&c_x\\
            0&f_y&c_y\\
            0&0&1\\
        \end{bmatrix}
\end{equation}
Distortion is classified as either radial or tangential, and the coefficients are solved for during the calibration process. The calibration is performed using a checkerboard of known size and the Matlab camera calibrator package~\cite{MATLAB:2022}. These parameters are used to correct for lens distortion and project camera points onto the world frame. The extrinsic parameters of the system are the locations of the sensors in the system with respect to a defined origin. 

In the stereo setup the intrinsics must be calculated for each of the cameras. In the case of the stereo setup, the extrinsic parameter is the transformation from one camera ($C_1$) to the other ($C_0$). This is expressed by a $4 \times 4$ matrix, $T_s$. A stereo camera setup is often defined by its \emph{baseline}, or the distance between the two cameras, which is the vector length of the translation from this matrix.
\begin{equation}
    T_s = \left[\begin{array}{ccc|c}
        & & &   \\
        & R_{C_1}^{C_0} & & t_{C_1}^{C_0} \\
        & & &  \\
        0 & 0 & 0 & 1
    \end{array}\right]
\end{equation}
where $R_{C_1}^{C_0}$ is the rotation matrix from $C_1$ to $C_0$ and $t_{C_1}^{C_0}$ is the translation from $C_1$ to $C_0$.
We conducted the extrinsic calibration using a checkerboard and the Matlab stereo camera calibrator package~\cite{MATLAB:2022}. To find the stereo baseline, the checkerboards are detected in each side of the stereo pair, and using the assumption that either the camera or the checkerboard is static, the pose of one camera in relation to the other is determined, giving us the transformation matrix $T_s$.

In the visual-inertial setup the extrinsic parameter is the transformation between the IMU and the camera ($C$). The extrinsic parameters are determined with an AprilTag board and the Kalibr toolbox~\cite{kalibr}. 

\begin{equation}
    T_s = \left[\begin{array}{ccc|c}
        & & &   \\
        & R_{\rm imu}^C  & & t_{\rm imu}^C \\
        & & &  \\
        0 & 0 & 0 & 1
    \end{array}\right]
\end{equation}
where $R_{\rm imu}^C$ is the $3 \times 3$ rotation matrix between the IMU and the camera $C$ and $t_{\rm imu}^C$ is the $3 \times 1$ translation matrix between the IMU and the camera $C$.

\section{Datasets}
\begin{figure*}[t!]
    \centering
    \includegraphics[width=\linewidth]{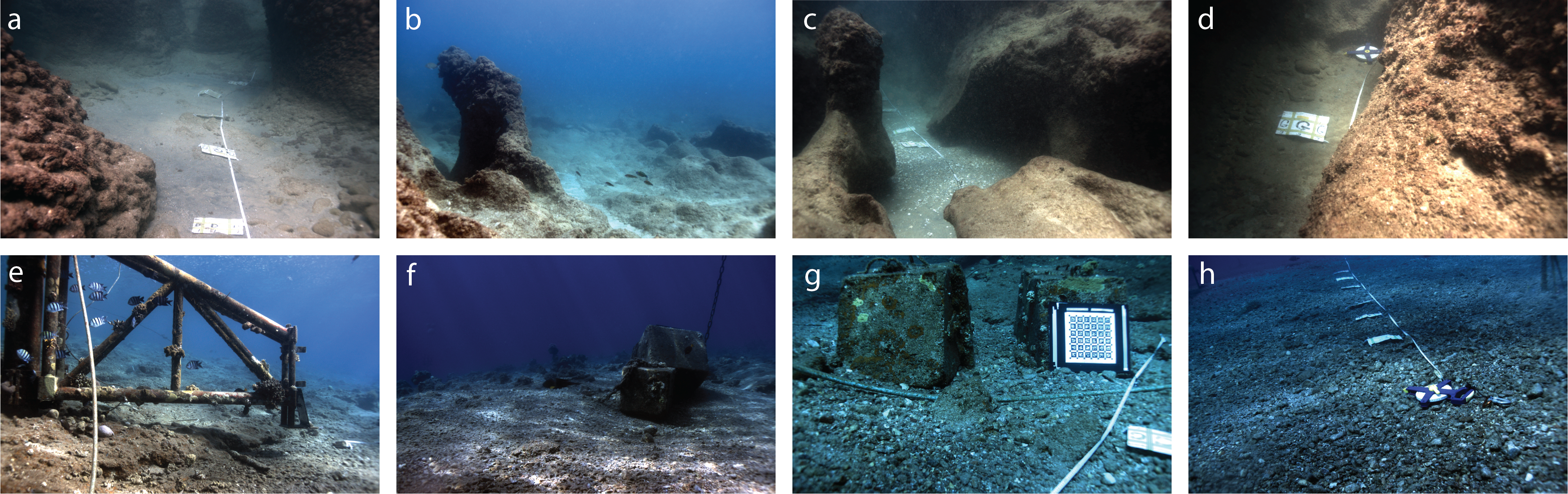}
    \caption{Examples from the visual-inertial datasets. \textbf{a)} U Canyon, \textbf{b)} Flatiron, \textbf{c)} Horse Canyon, \textbf{d)} Tiny Canyon, \textbf{e)} Northeast Path, \textbf{f)} Big Dice Loop, \textbf{g)} Landward Path, \textbf{h)} Cross Pyramid Loop.}
    \label{fig:dataset_examples_vi}
\end{figure*}

\begin{figure*}[t!]
    \centering
    \includegraphics[width=\linewidth]{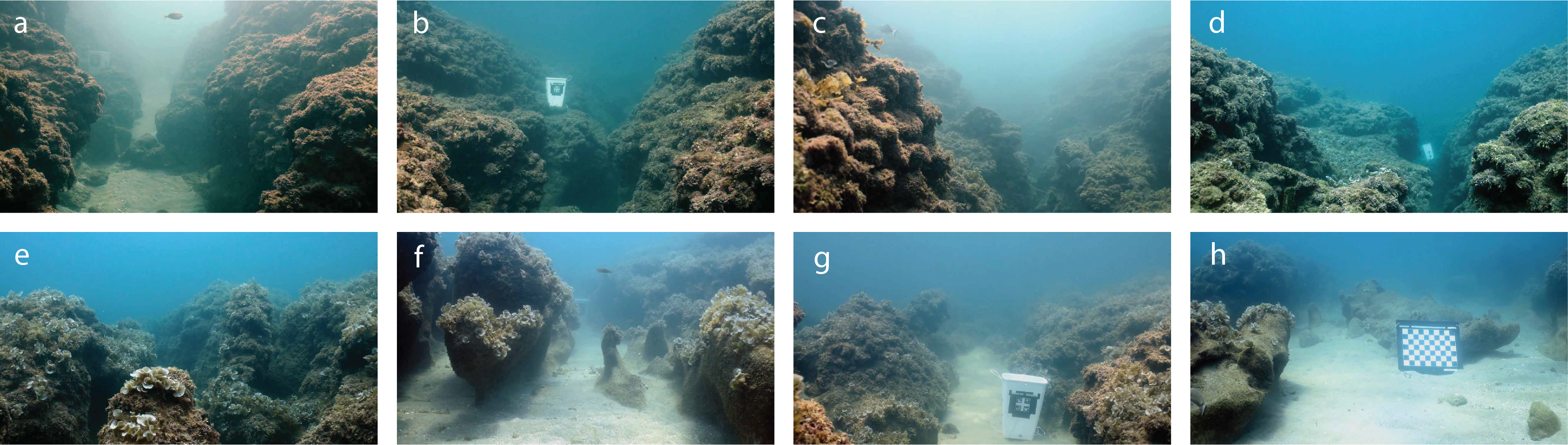}
    \caption{Examples from the stereo datasets \textbf{a-d)} Canyon 1, \textbf{e-g)} Rock Garden 1, \textbf{h)} Rock Garden 2}
    \label{fig:dataset_examples_stereo}
\end{figure*}

We present two types of datasets, stereo and visual-inertial. There are 12 visual-inertial datasets and 4 stereo datasets. The datasets are summarized in Table~\ref{datasets}. Examples of the image data can be seen in Figures~\ref{fig:dataset_examples_vi} and~\ref{fig:dataset_examples_stereo}. They were collected in the Mediterranean and Red Sea, on 8 dives. The stereo data includes images at 10~fps, with objects of known-size placed in the scene. On each dive, a calibration set was collected, used to determine the intrinsic and extrinsic (stereo baseline) parameters. The visual-inertial data includes images at 10~fps, also with objects of known-size placed in the scene, and IMU data at 20-100~Hz. A calibration set was also collected for each visual-inertial dive, allowing for calculation of the intrinsic and extrinsic (camera-IMU transformation) parameters. The datasets include the original images which are unenhanced and a second version enhanced with SeaErra software~\cite{seaerra}. 

\begin{table}[!h]
\small\sf
\caption{Datasets. \label{datasets}}
\begin{tabular}{p{2cm} p{3cm} p{2.2cm}} 
\toprule
\multicolumn{3}{c}{Visual-Inertial}\\
\midrule
 \textbf{Location}& \textbf{Dataset name} & \textbf{Image number} \\
 \multirow{4}{*}{Canyons} & U Canyon & 2895\\
 & Flatiron & 2475 \\
 & Horse Canyon & 2230\\
 & Tiny Canyon & 1012 \\
 \bottomrule
 \multirow{8}{*}{Red Sea} & Northeast Path & 2593\\ 
 & Landward Path & 1204\\
 & Dice Path & 1428\\
 & Pier Path & 1695\\
 & Coral Table Loop & 1017\\
 & Cross Pyramid Loop & 1652\\
 & Big Dice Loop & 3159\\
 & Sub Pier & 1091\\
\bottomrule
\textbf{Total} & & \textbf{22451}\\
\bottomrule
\end{tabular}\\[10pt]
\begin{tabular}{p{2cm} p{3cm} p{2.2cm}}
\toprule
\multicolumn{3}{c}{Stereo} \\
\midrule
 \textbf{Location}& \textbf{Dataset name} & \textbf{Image number} \\
 \multirow{2}{*}{Canyons} & Canyon 1 & 7606 L/R pairs \\
 & Canyon 2 & 2363 L/R pairs \\
\multirow{2}{*}{Shallow} & Flats & 2702 L/R pairs \\
& Rock Garden 1 & 5688 L/R pairs \\
& Rock Garden 2 & 1238 L/R pairs \\
\bottomrule
\textbf{Total} & & \textbf{19596}\\
\bottomrule
\end{tabular}\\[10pt]
\end{table}

\subsection{Visual-inertial data}

The visual-inertial datasets were captured in a few different locations, and are organized according to their environment type. The ``Canyons'' datasets were captured in Nachsholim, the Mediterranean, Israel, in three different canyons, totaling four datasets. The water depth for these four datasets range from 4-7~m. The ``Red sea'' datasets were captured in Eilat, Israel, in a few different areas, totaling eight datasets. The water depth for these eight datasets range from 3-8~m.  The visual-inertial data includes RGB images at 10~fps and IMU data taken at 20~Hz (``Canyons'') and 100~Hz (``Red Sea''). Each of the datasets include measuring tape laid on the seafloor, with visual targets of known size spaced a meter apart along the measuring tape as well as the calibration checkerboard or AprilTag board placed somewhere in the scene. Some datasets include ``loop closure,'' and some of the datasets include the same objects or similar trajectory as other datasets. The canyon datasets were collected in an area with substantial natural 3D structure, which serves as a challenging test for navigation and obstacle avoidance in complex environments. There is no man-made structure in the canyon datasets. The only objects of known size are the ones placed in the scene as mentioned earlier. More information about these objects can be found in the supplemental material. 

\begin{figure}[t]
    \centering
    \includegraphics[width=\linewidth]{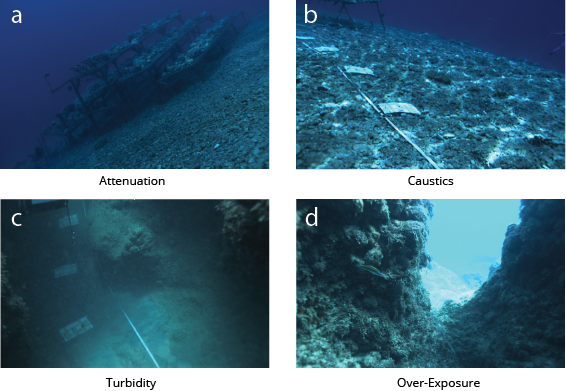}
    \caption{Underwater image phenomena}
    \label{fig:img_artifacts}
\end{figure}

There are a few image artifacts present in these datasets that are representative of the common challenges faced when imaging underwater. For a few of these sequences, the exposure was set to a static value, meaning that when there is a transition from low ambient light (in the canyon) to higher ambient light (out of the canyon, more shallow), some of the image frames are over exposed. While not ideal, this is an issue that is encountered when imaging in complex underwater environments with a system that does not have artificial lighting and fixed exposure. Underwater caustics are present in many of the datasets. Underwater caustics can be observed in shallow water, and are the result of a wavy sea-surface, which reflects or refracts light rays~\cite{caustics}, causing ever-changing light patterns on the seafloor. All of the datasets also exhibit attenuation and turbidity, affecting the range at which scene objects can be seen and the clarity at which we see them. Examples of all of these phenomena can be seen in Figure \ref{fig:img_artifacts}. 

\subsubsection{Visual-inertial data collection method.}
The visual-inertial datasets were collected with the BlueROV2.  To ensure a smooth route, the thrusters were shut off and the diver manually maneuvered the ROV throughout the environment. The camera and IMU were controlled by an operator on topside.  This means that the topside operator monitored the video feed and IMU measurements and adjusted camera settings throughout the dive. At the beginning of each dive a calibration set was collected, where the calibration board was positioned and intrinsic and extrinsic sets were collected. After the calibration set, the measuring tape and visual targets spaced a meter apart were laid on the seafloor and data collection commenced. Each dataset begins at the measuring tape. The diver then swims the ROV around the environment at a slow pace which mimics that of a small underwater vehicle.

\subsubsection{Visual-inertial dataset format.}
The visual-inertial datasets are organized as follows. 

\vspace{5mm}

\dirtree{%
.1 \textbf{\textit{location\textunderscore name}}.
.2 calibration.
.2 \textbf{\textit{dataset\textunderscore name}}.
}

\dirtree{%
.1 .
.2 .
.3 depth.
.4 \textbf{\textit{timestamp}}\textunderscore SeaErra\textunderscore abs\textunderscore depth.tif.
.4 $\dots$.
.3 imgs.
.4 \textbf{\textit{timestamp}}.tiff.
.4 $\dots$.
.3 seaErra.
.4 \textbf{\textit{timestamp}}\textunderscore SeaErra.tiff.
.4 $\dots$.
.3 imu.txt.
.3 notes.txt.
}

\vspace{5mm}

Each dataset includes the original images, enhanced  images, an ``imu.txt,'' file, depth maps for each image frames, model exports which contain camera poses and XYZ points, video files displaying different elements of the dataset and a ``notes.txt,'' file which contains details about the dataset.

\subsection{Stereo data}
The stereo datasets were captured on a single day, over the course of two dives in Nachsholim, Israel. The water depth in these datasets ranges from 3-8~m. There are two datasets captured within a canyon, at a maximum water depth of 8~m and two datasets captured on a flat shallower area at a maximum water depth of 5~m. The datasets include RGB stereo images collected at 10fps. Each dataset includes objects of known size, spaced evenly throughout the scene. Most of the datasets include loop closure events, meaning that locations were revisited throughout the dataset. There is also some challenging camera motion in these datasets, such as fast rotation. Other than the objects that were placed in the scene, there is no man-made structure in these scenes. The canyon datasets include plenty of natural structure in every frame. The flat datasets, however, have limited 3D structure and lots of homogeneous sandy areas. There is more turbidity than in the canyon datasets.

\subsubsection{Stereo dataset collection method.}
The stereo datasets were collected on the stereo rig in video mode. At the beginning of each dive, the exposure and focus was set, a calibration was performed and 3D objects of known size were placed throughout the scene. After the setup, data collection commenced by swimming through the environment with the stereo rig. During data collection, a point was made to revisit known locations, providing oppportunity for ``loop closure'' events.

\subsubsection{Stereo dataset format.}
The stereo datasets are organized as follows. 

\vspace{5mm}

\dirtree{%
.1 \textbf{\textit{location\textunderscore name}}.
.2 calibration.
.2 \textbf{\textit{dataset\textunderscore name}}.
.3 depth.
.4 LFT.
.5 LFT\textunderscore \textbf{\textit{image\textunderscore number}}\textunderscore abs\textunderscore depth.tif.
.5 $\dots$.
.4 RGT.
.5 RGT\textunderscore \textbf{\textit{image\textunderscore number}}\textunderscore abs\textunderscore depth.tif.
.5 $\dots$.
.3 imgs.
.4 LFT.
.5 LFT\textunderscore\textbf{\textit{image\textunderscore number}}.tiff.
.5 $\dots$.
.4 RGT.
.5 RGT\textunderscore \textbf{\textit{image\textunderscore number}}.tiff.
.5 $\dots$.
.3 seaErra.
.4 LFT.
.5 LFT\textunderscore \textbf{\textit{image\textunderscore number}}\textunderscore SeaErra.tiff.
.5 $\dots$.
.4 RGT.
.5 RGT\textunderscore \textbf{\textit{image\textunderscore number}}\textunderscore SeaErra.tiff.
.5 $\dots$.
.3 notes.txt.
}
\vspace{5mm}

\section{Groundtruth}
\begin{figure}[t]
    \centering
    \includegraphics[width=\linewidth]{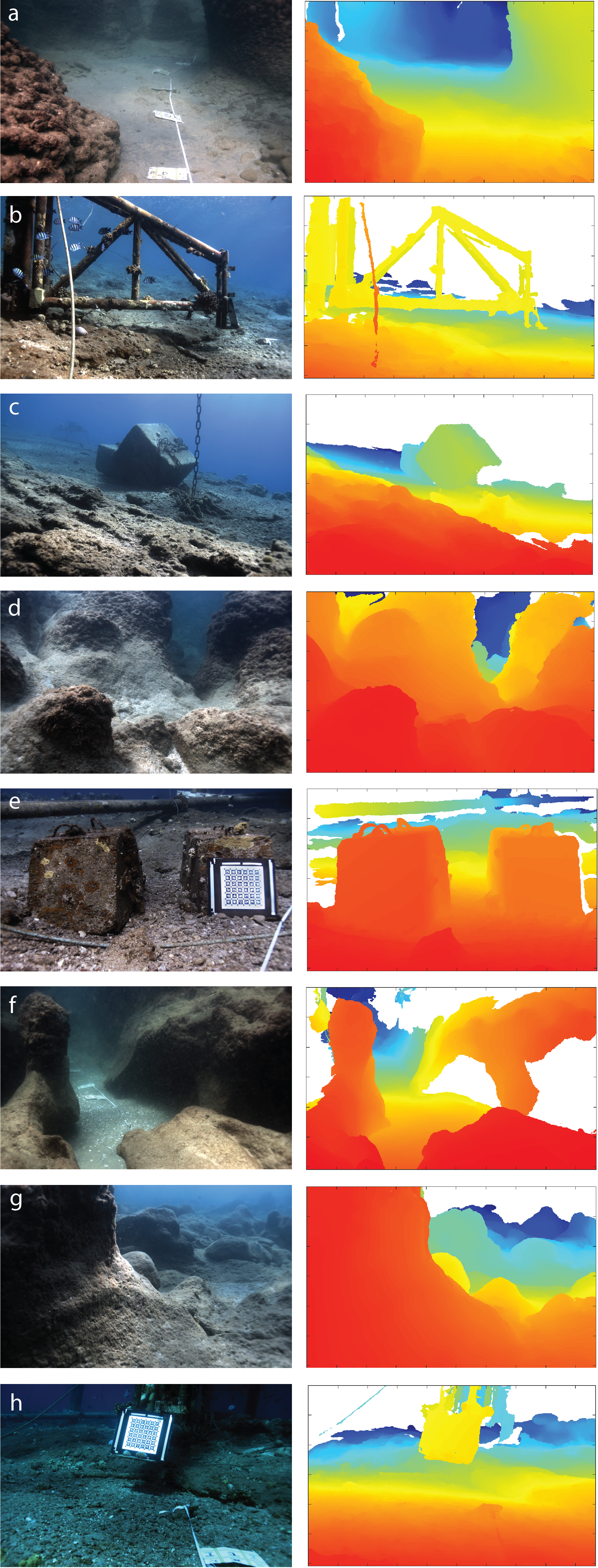}
    \caption{Ground truth examples from \textbf{a)} U Canyon, \textbf{b)} Northeast Path, \textbf{c)} Big Dice Loop, \textbf{d)} Horse Canyon, \textbf{e)} Landward Path \textbf{f)} Horse Canyon, \textbf{g)} Flatiron, \textbf{h)} Northeast Path. The white areas are where depth was not resolved. In this set of examples, depth ranges from 0~m at its closest to 12~m at its furthest.}
    \label{fig:ground_truth}
\end{figure}

Obtaining groundtruth underwater is a notorious challenge. GPS, a method for determining absolute position is not available and traditional methods for collecting 3D data such as LiDAR is limited underwater due to scattering and attenuation. So, we use Agisoft Metashape~\cite{agisoft} to generate per frame scaled depth maps offline from images and scale cues placed in the scene. Metashape uses SFM to estimate the camera poses and 3D structure of the scene given a sequence of multiple images. The software works by finding feature points in each image and matching them across images to create ``tie points.'' The output of this step is a tie point cloud and a camera pose for each image. At this stage, the scale references are identified and set, scaling the model to actual scene scale. Sparse depth maps are also generated at this time. The next step is to generate a dense point cloud from the camera poses and input images. The final step is to create the mesh model from the dense point cloud, from which we extract dense depth maps for each image frame as seen in Figure~\ref{fig:ground_truth}. The depth maps are then validated using a few methods. One method is to check known measurements of known distances in the scene. Another method we used for validation was by visual inspection. In other words, the depth maps are inspected by overlaying them on the input images, to check that objects represented in the depth maps align with the actual objects in the scene. The final validation is to compute the absolute error for objects of known size with AprilTags of known dimensions on them in the depth maps. The error found in this analysis was consistently less than 0.5~cm. It must be noted here that unfortunately these objects not present in all images, therefore we could only quantitatively check a small subset of the ground truth. The ground truth portion of the datasets includes depth maps, camera poses, and measurements of some objects in the scene (supplementary material, Figures \ref{fig:pier_measurements}-\ref{fig:scale_card}).

\section{Known issues}
There are a few known issues in the datasets that we would like to draw attention to for data users. Firstly, in the visual-inertial sets, a few of the images are over exposed. The over-exposure is present when there were drastic changes in light, for example between inside a canyon and out. This was resolved in later datasets. Another issue in the canyon datasets is that the IMU was set to collect inertial measurements at 20~Hz. Because there is no hardware synchronization between the IMU and the camera, and the camera collects images at 10~Hz, this is not ideal for visual-inertial applications. To solve this, the inertial measurements were interpolated to 100~Hz and in the red sea datasets the IMU was set to collect measurements at 100~Hz. 

Another known issue is in the ground truth section of the dataset. Being that we create the ground truth with SFM, a photogrammetry method, it is still prone to the negative effects of imaging underwater. This causes slight imperfections in the depth maps especially with objects in the far background. This is best visualized in the $<$dataset\textunderscore name$>$\textunderscore overlay.avi video.

\section{Summary}
In summary, we present the public with forward-looking stereo and visual-inertial underwater datasets. They are complete with either stereo image pairs or monocular images and inertial measurements, enhanced images, intrinsic and extrinsic calibrations, ground truth scaled depth maps and camera poses and measurements of known objects in the scene. We hope these datasets will be used to further the development of underwater autonomous systems, with the goal of working in complex environments close to potential obstacles, where a forward-looking view is necessary. We imagine that these datasets can be useful for testing SLAM and VIO algorithms and deep learning networks for depth estimation, for example. With the addition of depth maps in our datasets, these methods can be evaluated and compared, which is useful for the underwater robotics community.

\begin{acks}
The research was funded by Israel Science Foundation grant $\#680/18$, the Israeli Ministry of Science and Technology grant $\#3-15621$, the Israel Data Science Initiative (IDSI) of the Council for Higher Education in Israel, the Data Science Research Center at the University of Haifa, and the European Union’s Horizon 2020 research and innovation programme under grant agreement No. GA 101016958. 
A big thank you goes to all from the VISEAON Marine Imaging Lab. Thank you to Ohad Inbar for assistance with the BlueROV2, Deborah Levy for her help with image enhancement, ground truth generation and data collection, to Naama Pearl for her help with monocular depth reconstruction and to Judith Fischer, Aviad Avni and Matan Yuval for their help in data collection. 
\end{acks}

\bibliographystyle{SageH}
\bibliography{ref.bib}

\begin{thebibliography}{49}
\providecommand{\natexlab}[1]{#1}
\providecommand{\url}[1]{\texttt{#1}}
\providecommand{\urlprefix}{URL }
\expandafter\ifx\csname urlstyle\endcsname\relax
  \providecommand{\doi}[1]{DOI:\discretionary{}{}{}#1}\else
  \providecommand{\doi}{DOI:\discretionary{}{}{}\begingroup
  \urlstyle{rm}\Url}\fi

\bibitem[{Agarwal and Arora(2022)}]{pixelformer}
Agarwal A and Arora C (2022) Attention attention everywhere: Monocular depth
  prediction with skip attention.
\newblock \doi{10.48550/ARXIV.2210.09071}.
\newblock \urlprefix\url{https://arxiv.org/abs/2210.09071}.

\bibitem[{Agarwal et~al.(2011)Agarwal, Furukawa, Snavely, Simon, Curless, Seitz
  and Szeliski}]{rome_in_a_day}
Agarwal S, Furukawa Y, Snavely N, Simon I, Curless B, Seitz SM and Szeliski R
  (2011) Building rome in a day.
\newblock \emph{Commun. ACM} 54(10): 105–112.
\newblock \doi{10.1145/2001269.2001293}.
\newblock
  \urlprefix\url{https://doi-org.ezproxy.haifa.ac.il/10.1145/2001269.2001293}.

\bibitem[{Agisoft(2018)}]{agisoft}
Agisoft (2018) {Metashape PhotoScan Professional (Software)}.
\newblock \emph{Retrieved from http://www.agisoft.com/downloads/installer/.}
  Version 1.4.5.

\bibitem[{Agrafiotis et~al.(2018)Agrafiotis, Skarlatos, Forbes, Poullis,
  Skamantzari and Georgopoulos}]{caustics}
Agrafiotis P, Skarlatos D, Forbes T, Poullis C, Skamantzari M and Georgopoulos
  A (2018) Underwater photogrammetry in very shallow waters: Main challenges
  and caustics effect removal.
\newblock \emph{The International Archives of the Photogrammetry, Remote
  Sensing and Spatial Information Sciences} XLII-2.

\bibitem[{Akkaynak and Treibitz(2018)}]{derya}
Akkaynak D and Treibitz T (2018) A revised underwater image formation model.
\newblock \emph{2018 IEEE/CVF Conference on Computer Vision and Pattern
  Recognition} : 6723--6732.

\bibitem[{Bloesch et~al.(2017)Bloesch, Burri, Omari, Hutter and
  Siegwart}]{rovio}
Bloesch M, Burri M, Omari S, Hutter M and Siegwart R (2017) Iterated extended
  kalman filter based visual-inertial odometry using direct photometric
  feedback.
\newblock \emph{The International Journal of Robotics Research} 36(10):
  1053--1072.
\newblock \doi{10.1177/0278364917728574}.
\newblock \urlprefix\url{https://doi.org/10.1177/0278364917728574}.

\bibitem[{{Blue Robotics}(2022)}]{bluerov2}
{Blue Robotics} (2022) {BlueROV2 - affordable and capable underwater ROV}.
\newblock \urlprefix\url{https://bluerobotics.com/store/rov/bluerov2/}.

\bibitem[{Burri et~al.(2016)Burri, Nikolic, Gohl, Schneider, Rehder, Omari,
  Achtelik and Siegwart}]{euroc}
Burri M, Nikolic J, Gohl P, Schneider T, Rehder J, Omari S, Achtelik MW and
  Siegwart R (2016) The euroc micro aerial vehicle datasets.
\newblock \emph{The International Journal of Robotics Research} 35(10):
  1157--1163.
\newblock \doi{10.1177/0278364915620033}.
\newblock \urlprefix\url{https://doi.org/10.1177/0278364915620033}.

\bibitem[{Campos et~al.(2021)Campos, Elvira, Rodríguez, M.~Montiel and
  D.~Tardós}]{orb-slam3}
Campos C, Elvira R, Rodríguez JJG, M~Montiel JM and D~Tardós J (2021)
  Orb-slam3: An accurate open-source library for visual, visual–inertial, and
  multimap slam.
\newblock \emph{IEEE Transactions on Robotics} 37(6): 1874--1890.

\bibitem[{Carreras et~al.(2018)Carreras, Hernández, Vidal, Palomeras, Ribas
  and Ridao}]{sparus_ii}
Carreras M, Hernández JD, Vidal E, Palomeras N, Ribas D and Ridao P (2018)
  Sparus ii auv—a hovering vehicle for seabed inspection.
\newblock \emph{IEEE Journal of Oceanic Engineering} 43(2): 344--355.
\newblock \doi{10.1109/JOE.2018.2792278}.

\bibitem[{Engel et~al.(2016)Engel, Koltun and Cremers}]{dso}
Engel J, Koltun V and Cremers D (2016) Direct sparse odometry.
\newblock \emph{CoRR} abs/1607.02565.
\newblock \urlprefix\url{http://arxiv.org/abs/1607.02565}.

\bibitem[{Engel et~al.(2014)Engel, Sch{\"o}ps and Cremers}]{lsd-slam}
Engel J, Sch{\"o}ps T and Cremers D (2014) Lsd-slam: Large-scale direct
  monocular slam.
\newblock In: Fleet D, Pajdla T, Schiele B and Tuytelaars T (eds.)
  \emph{Computer Vision -- ECCV 2014}. Cham: Springer International Publishing,
  pp. 834--849.

\bibitem[{Ferrera et~al.(2019)Ferrera, Creuze, Moras and
  Trouv\'e-Peloux}]{aqualoc}
Ferrera M, Creuze V, Moras J and Trouv\'e-Peloux P (2019) Aqualoc: An
  underwater dataset for visual–inertial–pressure localization.
\newblock \emph{The International Journal of Robotics Research} 38(14):
  1549–1559.

\bibitem[{Forster et~al.(2014)Forster, Pizzoli and Scaramuzza}]{svo}
Forster C, Pizzoli M and Scaramuzza D (2014) Svo: Fast semi-direct monocular
  visual odometry.
\newblock In: \emph{2014 IEEE International Conference on Robotics and
  Automation (ICRA)}. pp. 15--22.
\newblock \doi{10.1109/ICRA.2014.6906584}.

\bibitem[{Geiger et~al.(2013)Geiger, Lenz, Stiller and Urtasun}]{kitti}
Geiger A, Lenz P, Stiller C and Urtasun R (2013) Vision meets robotics: The
  kitti dataset.
\newblock \emph{International Journal of Robotics Research (IJRR)} .

\bibitem[{Hartley and Zisserman(2004)}]{Hartley2004}
Hartley RI and Zisserman A (2004) \emph{Multiple View Geometry in Computer
  Vision}.
\newblock Second edition. Cambridge University Press, ISBN: 0521540518.

\bibitem[{Huai and Huang(2022)}]{r-vio}
Huai Z and Huang G (2022) Robocentric visual-inertial odometry.
\newblock \emph{The International Journal of Robotics Research} 41(7):
  667--689.

\bibitem[{Hugyfot(2022)}]{hugyfot}
Hugyfot (2022) Hfn-d810: Nikon d810 underwater housing.
\newblock \urlprefix\url{https://www.hugyfot.com/l}.

\bibitem[{{iDS}(2022)}]{ids-cam}
{iDS} (2022) Ui-3260cp rev. 2.
\newblock \urlprefix\url{https://www.ids-imaging.us/}.

\bibitem[{Jaffe(1990)}]{jaffe}
Jaffe J (1990) Computer modeling and the design of optimal underwater imaging
  systems.
\newblock \emph{IEEE Journal of Oceanic Engineering} 15(2): 101--111.
\newblock \doi{10.1109/48.50695}.

\bibitem[{Kim et~al.(2022)Kim, Ga, Ahn, Joo, Chun and Kim}]{glpdepth}
Kim D, Ga W, Ahn P, Joo D, Chun S and Kim J (2022) Global-local path networks
  for monocular depth estimation with vertical cutdepth.
\newblock \emph{CoRR} abs/2201.07436.
\newblock \urlprefix\url{https://arxiv.org/abs/2201.07436}.

\bibitem[{Li and Snavely(2018)}]{megadepth}
Li Z and Snavely N (2018) Megadepth: Learning single-view depth prediction from
  internet photos.
\newblock In: \emph{Proceedings of the IEEE Conference on Computer Vision and
  Pattern Recognition (CVPR)}.

\bibitem[{Li et~al.(2022)Li, Wang, Liu and Jiang}]{binsformer}
Li Z, Wang X, Liu X and Jiang J (2022) Binsformer: Revisiting adaptive bins for
  monocular depth estimation.
\newblock \doi{10.48550/ARXIV.2204.00987}.
\newblock \urlprefix\url{https://arxiv.org/abs/2204.00987}.

\bibitem[{Lindenberger et~al.(2021)Lindenberger, Sarlin, Larsson and
  Pollefeys}]{pixel_perfect}
Lindenberger P, Sarlin P, Larsson V and Pollefeys M (2021) Pixel-perfect
  structure-from-motion with featuremetric refinement.
\newblock \emph{CoRR} abs/2108.08291.
\newblock \urlprefix\url{https://arxiv.org/abs/2108.08291}.

\bibitem[{Longuet-Higgins(1987)}]{sfm_og}
Longuet-Higgins H (1987) A computer algorithm for reconstructing a scene from
  two projections.
\newblock In: Fischler MA and Firschein O (eds.) \emph{Readings in Computer
  Vision}. San Francisco (CA): Morgan Kaufmann.
\newblock ISBN 978-0-08-051581-6, pp. 61--62.
\newblock \doi{https://doi.org/10.1016/B978-0-08-051581-6.50012-X}.
\newblock
  \urlprefix\url{https://www.sciencedirect.com/science/article/pii/B978008051581650012X}.

\bibitem[{Luczynski et~al.(2021)Luczynski, Willners, Vargas, Roe, Xu, Cao,
  Petillot and Wang}]{inspection_intervention_dataset}
Luczynski T, Willners JS, Vargas E, Roe J, Xu S, Cao Y, Petillot YR and Wang S
  (2021) Underwater inspection and intervention dataset.
\newblock \emph{CoRR} abs/2107.13628.

\bibitem[{Mallios et~al.(2017)Mallios, Vidal, Campos and
  Carreras}]{underwater_caves}
Mallios A, Vidal E, Campos R and Carreras M (2017) Underwater caves sonar data
  set.
\newblock \emph{The International Journal of Robotics Research} 36(12):
  1247--1251.

\bibitem[{Mapillary(2021)}]{opensfm}
Mapillary (2021) {OpenSFM}.
\newblock \emph{Retrieved from https://opensfm.org/.} Version 0.5.1.

\bibitem[{MATLAB(2022)}]{MATLAB:2022}
MATLAB (2022) \emph{R2022a}.
\newblock Natick, Massachusetts: The MathWorks Inc.

\bibitem[{Mohr et~al.(1995)Mohr, Quan and Veillon}]{sfm_uncalibrated}
Mohr R, Quan L and Veillon F (1995) {Relative 3D Reconstruction Using Multiple
  Uncalibrated Images}.
\newblock \emph{{The International Journal of Robotics Research}} 14(6):
  619--632.
\newblock \doi{10.1177/027836499501400607}.
\newblock \urlprefix\url{https://hal.inria.fr/inria-00548396}.

\bibitem[{Moulon et~al.(2016)Moulon, Monasse, Perrot and Marlet}]{openmvg}
Moulon P, Monasse P, Perrot R and Marlet R (2016) Open{MVG}: Open multiple view
  geometry.
\newblock In: \emph{International Workshop on Reproducible Research in Pattern
  Recognition}. Springer, pp. 60--74.

\bibitem[{Mur-Artal et~al.(2015)Mur-Artal, Montiel and Tardós}]{orb-slam}
Mur-Artal R, Montiel JMM and Tardós JD (2015) Orb-slam: A versatile and
  accurate monocular slam system.
\newblock \emph{IEEE Transactions on Robotics} 31(5): 1147--1163.
\newblock \doi{10.1109/TRO.2015.2463671}.

\bibitem[{Mur-Artal and Tardós(2017)}]{mur-artal}
Mur-Artal R and Tardós JD (2017) Visual-inertial monocular slam with map
  reuse.
\newblock \emph{IEEE Robotics and Automation Letters} 2(2): 796--803.
\newblock \doi{10.1109/LRA.2017.2653359}.

\bibitem[{Nikon(2022)}]{nikond810}
Nikon (2022) Nikon d810: Full-frame dslr.
\newblock \urlprefix\url{https://www.nikonusa.com/l}.

\bibitem[{Qin et~al.(2019)Qin, Pan, Cao and Shen}]{vins-fusion}
Qin T, Pan J, Cao S and Shen S (2019) A general optimization-based framework
  for local odometry estimation with multiple sensors.

\bibitem[{Rahman et~al.(2018)Rahman, Quattrini~Li and Rekleitis}]{sCarolina}
Rahman S, Quattrini~Li A and Rekleitis I (2018) Sonar visual inertial slam of
  underwater structures.
\newblock \emph{2018 IEEE International Conference on Robotics and Automation
  (ICRA)} .

\bibitem[{Rehder et~al.(2016)Rehder, Nikolic, Schneider, Hinzmann and
  Siegwart}]{kalibr}
Rehder J, Nikolic J, Schneider T, Hinzmann T and Siegwart R (2016) Extending
  kalibr: Calibrating the extrinsics of multiple imus and of individual axes.
\newblock In: \emph{2016 IEEE International Conference on Robotics and
  Automation (ICRA)}. pp. 4304--4311.
\newblock \doi{10.1109/ICRA.2016.7487628}.

\bibitem[{Rosinol et~al.(2020)Rosinol, Abate, Chang and
  Carlone}]{Rosinol20icra-Kimera}
Rosinol A, Abate M, Chang Y and Carlone L (2020) Kimera: an open-source library
  for real-time metric-semantic localization and mapping.
\newblock In: \emph{IEEE Intl. Conf. on Robotics and Automation (ICRA)}.
\newblock \urlprefix\url{https://github.com/MIT-SPARK/Kimera}.

\bibitem[{Scaramuzza and Zhang(2019)}]{vio}
Scaramuzza D and Zhang Z (2019) Visual-inertial odometry of aerial robots.
\newblock \emph{CoRR} abs/1906.03289.
\newblock \urlprefix\url{http://arxiv.org/abs/1906.03289}.

\bibitem[{Schaffalitzky and Zisserman(2002)}]{vacation_sfm}
Schaffalitzky F and Zisserman A (2002) Multi-view matching for unordered image
  sets, or ``how do i organize my holiday snaps?''.
\newblock In: Heyden A, Sparr G, Nielsen M and Johansen P (eds.) \emph{Computer
  Vision --- ECCV 2002}. Berlin, Heidelberg: Springer Berlin Heidelberg.
\newblock ISBN 978-3-540-47969-7, pp. 414--431.

\bibitem[{Schneider et~al.(2017)Schneider, Dymczyk, Fehr, Egger, Lynen,
  Gilitschenski and Siegwart}]{maplab}
Schneider T, Dymczyk M, Fehr M, Egger K, Lynen S, Gilitschenski I and Siegwart
  R (2017) maplab: An open framework for research in visual-inertial mapping
  and localization.
\newblock \emph{CoRR} abs/1711.10250.
\newblock \urlprefix\url{http://arxiv.org/abs/1711.10250}.

\bibitem[{Sch\"{o}nberger and Frahm(2016)}]{colmap}
Sch\"{o}nberger JL and Frahm JM (2016) {Structure-from-Motion Revisited}.
\newblock In: \emph{Conference on Computer Vision and Pattern Recognition
  (CVPR)}.

\bibitem[{Sturm et~al.(2012)Sturm, Engelhard, Endres, Burgard and
  Cremers}]{tum-rgb-d}
Sturm J, Engelhard N, Endres F, Burgard W and Cremers D (2012) A benchmark for
  the evaluation of rgb-d slam systems .

\bibitem[{{The Phoblographer}(2014)}]{nikon_sketch}
{The Phoblographer} (2014) Nikon d800 sketches.
\newblock
  \urlprefix\url{https://www.thephoblographer.com/2014/03/25/reports-state-nikon-d800s-may-works/}.

\bibitem[{Treibitz et~al.(2022)Treibitz, Levy, Goldfracht, Akkaynak and
  Bekerman}]{seaerra}
Treibitz T, Levy D, Goldfracht Y, Akkaynak D and Bekerman Y (2022) Seaerra.
\newblock \emph{SeaErra} .

\bibitem[{{VectorNav}(2022)}]{vn-100}
{VectorNav} (2022) Vn-100 imu.
\newblock \urlprefix\url{https://www.vectornav.com/}.

\bibitem[{Wofk et~al.(2019)Wofk, Ma, Yang, Karaman and Sze}]{fastdepth}
Wofk D, Ma F, Yang TJ, Karaman S and Sze V (2019) Fastdepth: Fast monocular
  depth estimation on embedded systems.
\newblock In: \emph{2019 International Conference on Robotics and Automation
  (ICRA)}. pp. 6101--6108.
\newblock \doi{10.1109/ICRA.2019.8794182}.

\bibitem[{Yin et~al.(2020)Yin, Zhang, Wang, Niklaus, Mai, Chen and
  Shen}]{leres}
Yin W, Zhang J, Wang O, Niklaus S, Mai L, Chen S and Shen C (2020) Learning to
  recover 3d scene shape from a single image.
\newblock \emph{CoRR} abs/2012.09365.
\newblock \urlprefix\url{https://arxiv.org/abs/2012.09365}.

\bibitem[{Yuan et~al.(2022)Yuan, Gu, Dai, Zhu and Tan}]{newcrfs}
Yuan W, Gu X, Dai Z, Zhu S and Tan P (2022) New crfs: Neural window
  fully-connected crfs for monocular depth estimation.
\newblock \doi{10.48550/ARXIV.2203.01502}.
\newblock \urlprefix\url{https://arxiv.org/abs/2203.01502}.

\end{thebibliography}

\clearpage
\begin{sm}

\begin{figure}[h]
    \centering
    \includegraphics[width=\linewidth]{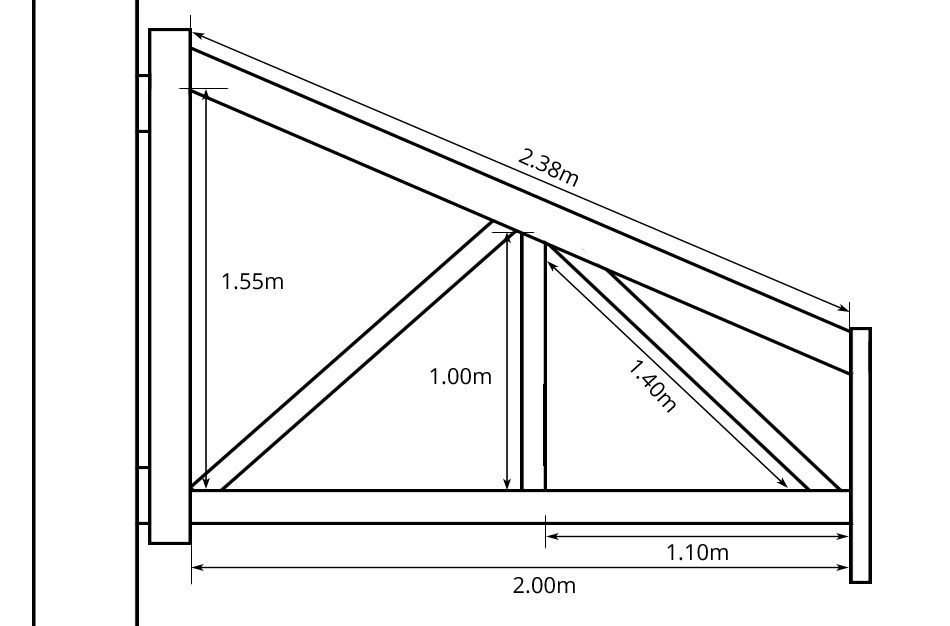}
    \caption{Pier measurements}
    \label{fig:pier_measurements}
\end{figure}

\begin{figure}[h]
    \includegraphics[width=0.9\linewidth]{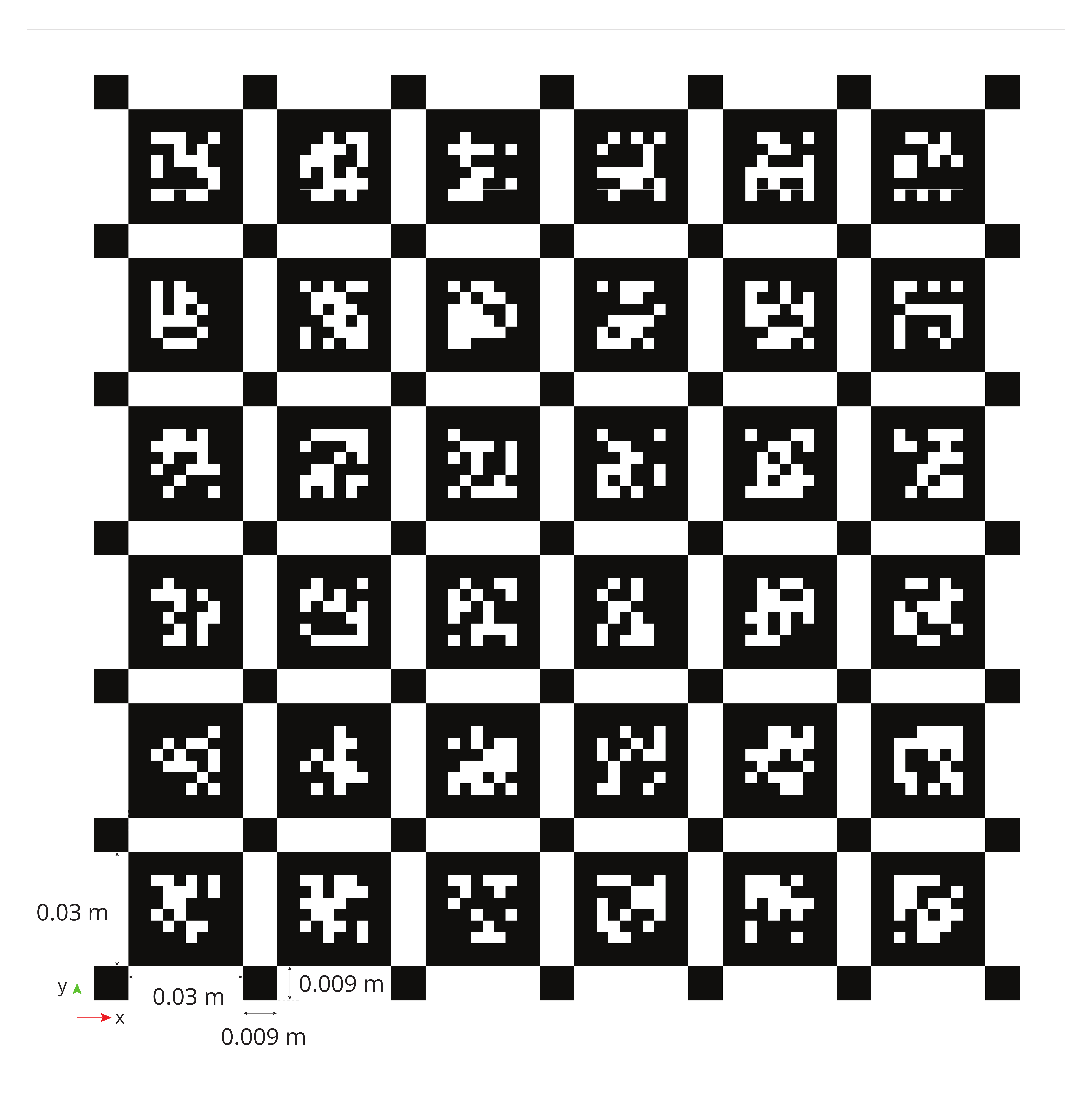}
    \caption{AprilTag board dimensions}
    \label{fig:april_board}
\end{figure}

\begin{figure}[h]
    \centering
    \includegraphics[width=\linewidth]{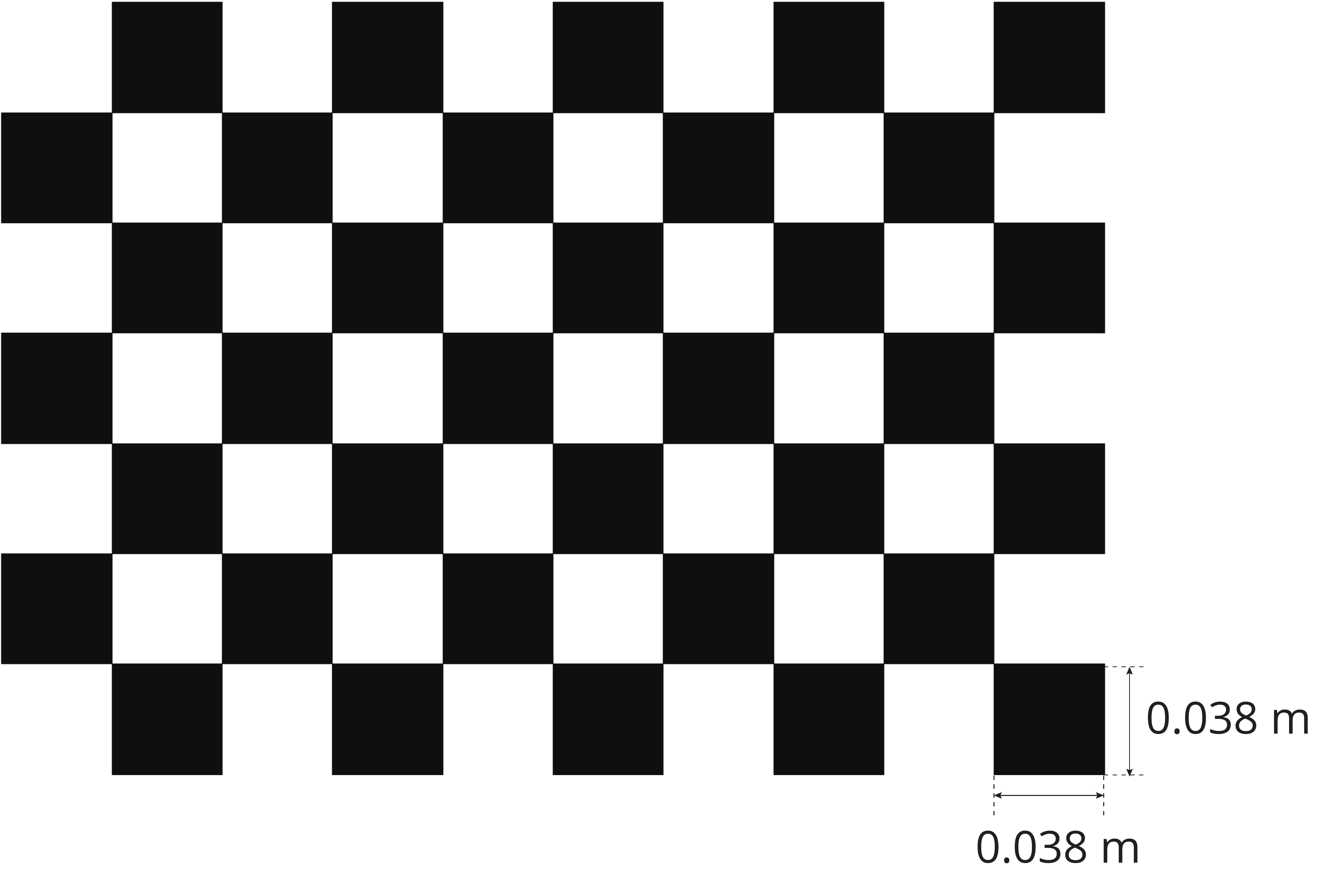}
    \caption{Checkerboard dimensions}
    \label{fig:checkerboard}
\end{figure}

\begin{figure}[h]
    \centering
    \includegraphics[width=\linewidth]{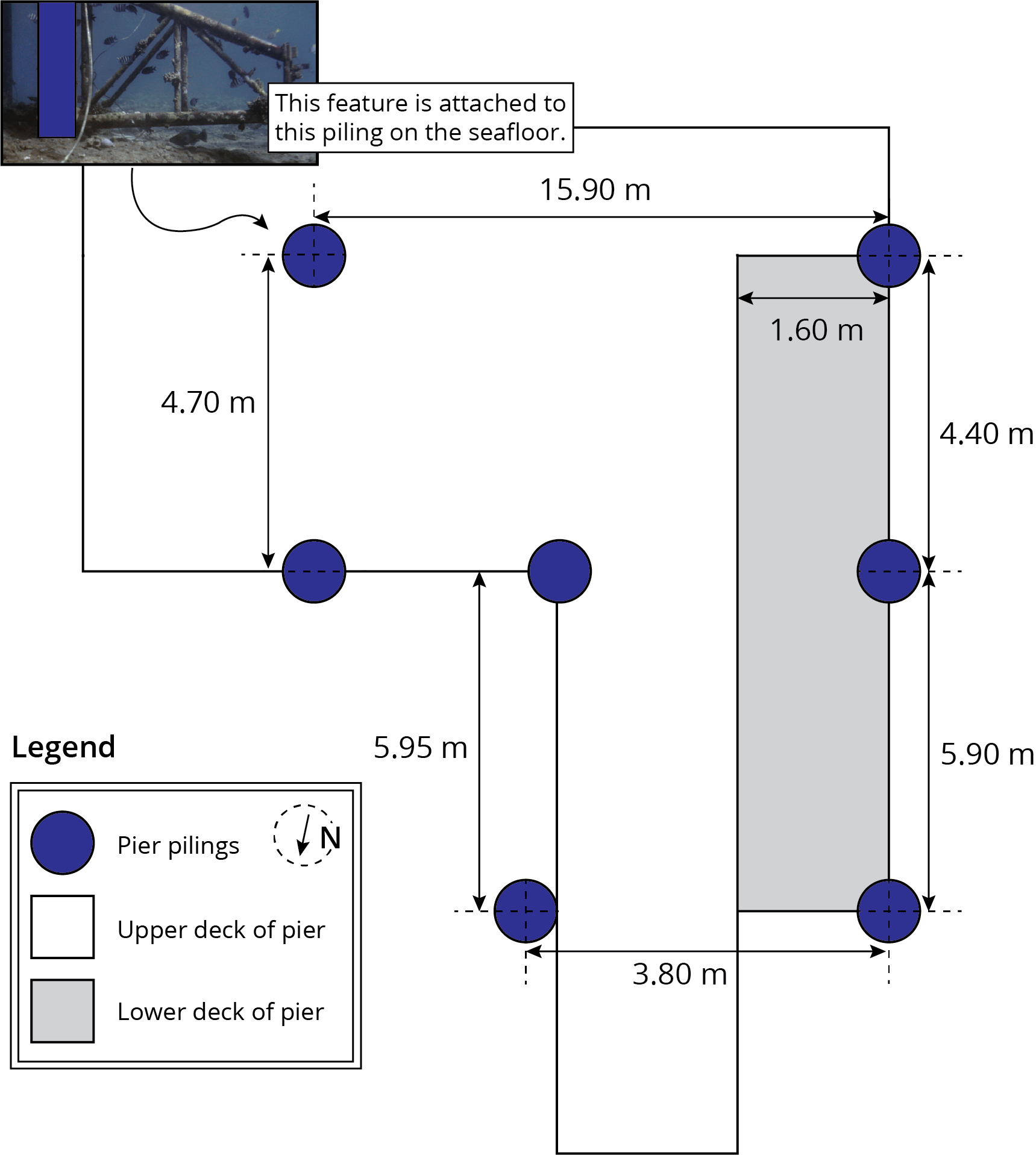}
    \caption{Drawing of pier with dimensions from above.}
    \label{fig:pier_birds}
\end{figure}

\begin{figure}[h]
    \centering
    \includegraphics[width=\linewidth]{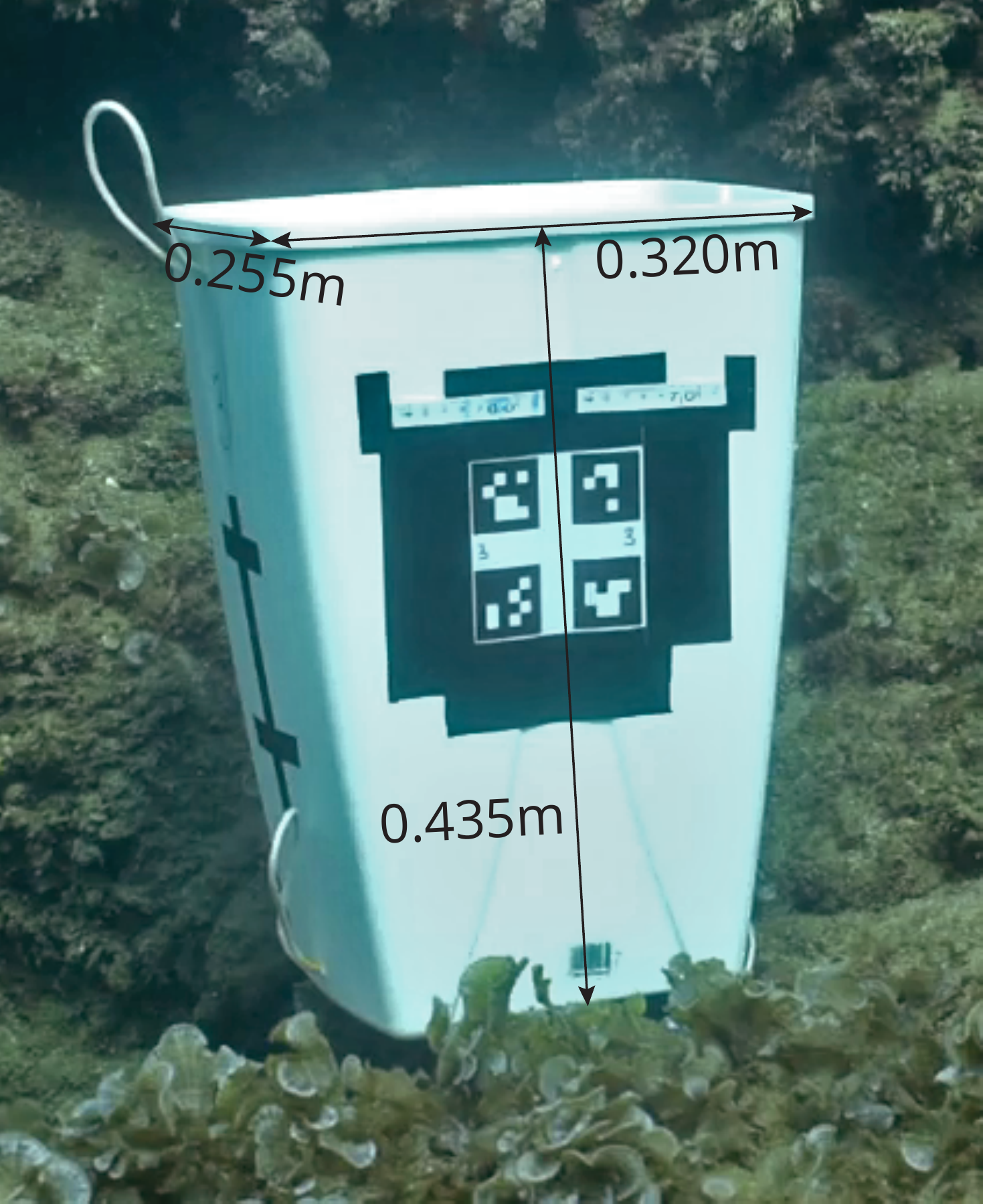}
    \caption{Dimensions of known object in stereo datasets.}
    \label{fig:trashcan}
\end{figure}

\begin{figure}[H]
    \centering
    \includegraphics[width=\linewidth]{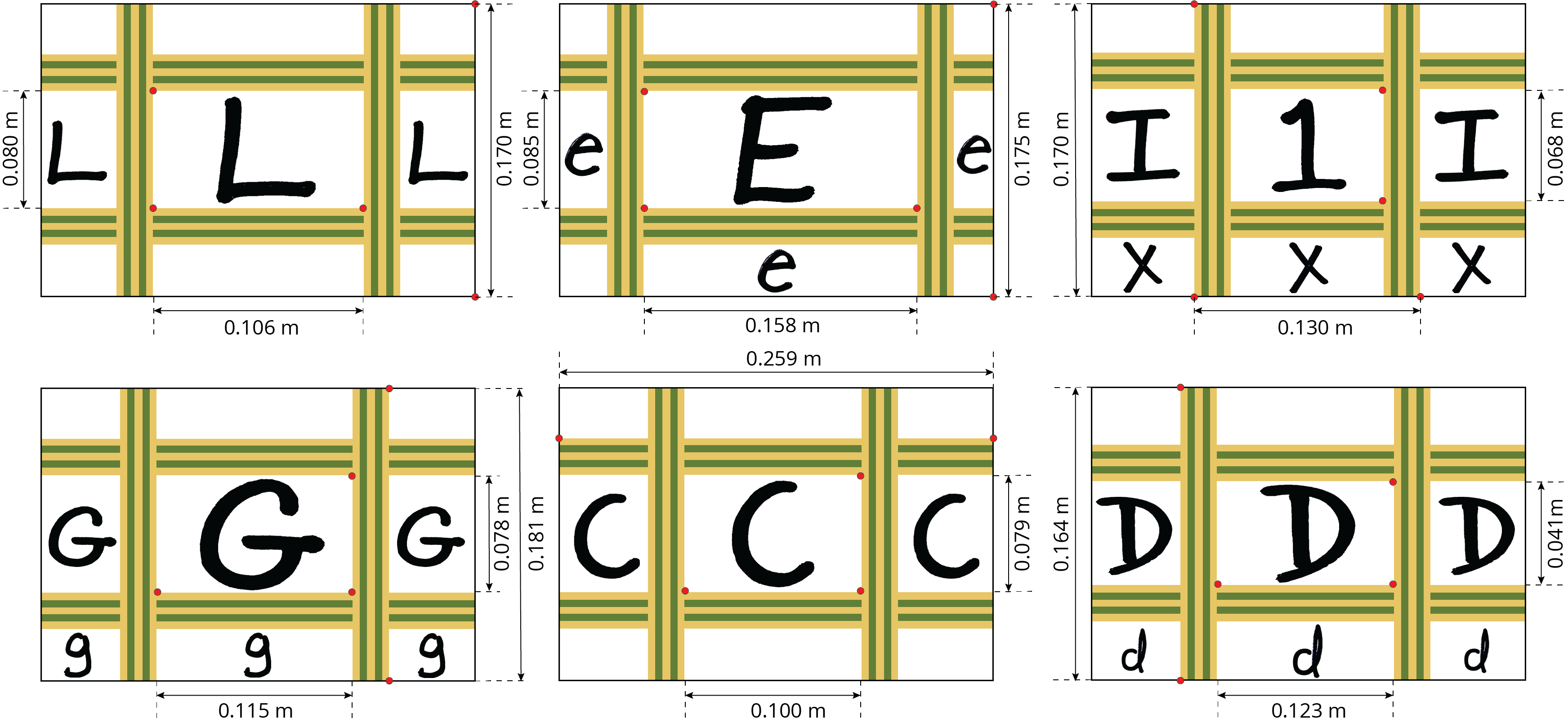}
    \caption{Scale card measurements}
    \label{fig:scale_card}
\end{figure}

\end{sm}
\end{document}